%% file: lrec2020W-xample-kc.tex
\documentclass[10pt, a4paper]{article}
\usepackage{lrec}
\usepackage{multibib}
\newcites{languageresource}{Language Resources}
\usepackage{graphicx}
\usepackage{tabularx}
\usepackage{soul}
\usepackage{epstopdf}
\usepackage[utf8]{inputenc}

\usepackage{hyperref}
\usepackage{xstring}
\usepackage{amsmath,graphicx}
\usepackage{color}
\usepackage{subcaption}

\title{Effects of Language Relatedness for Cross-lingual Transfer Learning in \\Character-Based Language Models}

\name{Mittul Singh$^*$, Peter Smit$^{*\ddagger}$, Sami Virpioja$^\dagger$, Mikko Kurimo$^*$}

\address{$^*$Department of Signal Processing and Acoustics, Aalto University, Espoo, Finland\\
        $^\dagger$Department of Digital Humanities, Helsinki University, Helsinki, Finland\\
        $^\ddagger$Inscripta, Helsinki, Finland\\
{\small \texttt{firstname.lastname@\{aalto,helsinki\}.fi}}}

\abstract{
\input{abstract.tex} \\ \newline \Keywords{Cross-lingual transfer, Character language models, Low-resource ASR}}

\begin{document}

\maketitleabstract

\input{introduction.tex}
\input{related_work.tex}
\input{data.tex}
\input{perplexity.tex}
\input{asr.tex}
\input{conclusion.tex}

\section{Bibliographical References}
\bibliographystyle{lrec}
\bibliography{asru}

\end{document}

%% file: abstract.tex
Character-based Neural Network Language Models (NNLM) have the advantage of smaller vocabulary and thus faster training times in comparison to NNLMs based on multi-character units. However, in low-resource scenarios, both the character and multi-character NNLMs suffer from data sparsity. In such scenarios, cross-lingual transfer has improved multi-character NNLM performance by allowing information transfer from a \textit{source} to the \textit{target} language. In the same vein, we propose to use cross-lingual transfer for character NNLMs applied to low-resource Automatic Speech Recognition (ASR). However, applying cross-lingual transfer to character NNLMs is not as straightforward. We observe that relatedness of the source language plays an important role in cross-lingual pretraining of character NNLMs. We evaluate this aspect on ASR tasks for two target languages: Finnish (with English and Estonian as source) and Swedish (with Danish, Norwegian, and English as source). Prior work has observed no difference between using the related or unrelated language for multi-character NNLMs. We, however, show that for character-based NNLMs, only pretraining with a related language improves the ASR performance, and using an unrelated language may deteriorate it. We also observe that the benefits are larger when there is much lesser target data than source data.

%% file: introduction.tex
\section{Introduction}
Multilingual training of language models has successfully leveraged datasets from other languages to improve Neural Network Language Modeling (NNLM) performance in low-resource scenarios \cite{DBLP:conf/acl/KimGN19,DBLP:conf/nips/ConneauL19,DBLP:journals/corr/abs-1911-02116,aharoni-etal-2019-massively}. 
One such method for training NNLM is the multi-task-based approach, where multiple language corpora train the model simultaneously \cite{aharoni-etal-2019-massively}. Another approach is cross-lingual pretraining, where the NNLM is trained on a set of \textit{source} languages followed by fine-tuning on the \textit{target} language \cite{DBLP:conf/acl/KimGN19,DBLP:conf/nips/ConneauL19,DBLP:journals/corr/abs-1911-02116}. The second approach, explored in this work, is favorable when re-training with the large source data is time-consuming as an existing trained source model's weights can be transferred to the target model and then fine-tuned on the smaller target data.

Cross-lingually pretrained NNLMs have utilized multi-character units to construct large shared vocabulary to allow the positive transfer of information from source to target. Instead of multi-character units, we explore a single character as a modeling unit for applying cross-lingual pretraining. This choice has the advantage of reducing the vocabulary size by several orders of magnitude and providing a larger intersection of vocabulary terms than multi-character units. In this paper, we apply cross-lingual pretraining to character NNLMs. However, this off-the-shelf application is not trivial. 
For multi-character based NNLMs, cross-lingual pretraining works by sharing information across various source languages independent of relatedness to the target language in terms of closeness in the language family tree\footnote{\url{https://en.wikipedia.org/wiki/Language_family}}. In contrast, for character-based NNLMs, a source language in the same family subtree as the target (related) affects the downstream performance positively than from an unrelated source language. 

We experiment with available Finnish and Swedish Automatic Speech Recognition (ASR) systems in a simulated low-resource ASR scenario by limiting the language modeling resources. We apply pretraining with two source languages (Estonian and English) for Finnish ASR and three source languages (Danish, English, and Norwegian) for Swedish ASR. In our experiments, we observe perplexity and ASR performance improvements when pretraining NNLMs with related languages (i.e. Estonian for Finnish and Danish and Norwegian for Swedish), whereas pretraining NNLMs on English performs adversely.

We also study the impact on cross-lingual transfer due to the target data size and number of source model layers transferred.  Relatively, smaller amounts of target language data than the source language data leads to more considerable ASR performance improvements. Moreover, we find that pretrained NNLMs perform best when we transfer only the parameters of the lowest layer of the source model.

%% file: related_work.tex
\section{Related Work}
In our work, we follow the cross-lingual pretraining scheme utilizing a shared vocabulary as proposed by Zhuang et al. \cite{DBLP:conf/interspeech/ZhuangGRPL17}, where they transfer all the hidden layers except the final layer from the source model to the target model. For NNLMs, such an application does not obtain the best results. In sections \ref{sec:ppl} and \ref{sec:asr}, we present results to support this observation.

Concurrently, Lample and Conneau \cite{DBLP:conf/nips/ConneauL19} have also shown that cross-lingual pretraining can improve the performance of language models on intrinsic measures like perplexity. They train a multi-character transformer-based language model with a masked language model training procedure for cross-lingual pretraining. In their model, multi-character units from both the source and target languages are combined to form one large vocabulary. This large shared vocabulary leads to a large output layer, which can be inefficient to train. The layer size can be reduced by shortlists and class-based models \cite{DBLP:conf/icassp/Goodman01,DBLP:conf/interspeech/LeOMAGY11}, or approximated by applying a hierarchical softmax \cite{DBLP:conf/aistats/MorinB05}. Instead, we choose characters as the basic unit of modeling, which provides a more natural way of reducing the vocabulary size. Simultaneously, this choice supports the cross-lingual information transfer by providing a larger intersection of vocabulary terms than multi-character units. 

For cross-lingual pretraining, language relatedness remains an unexplored factor, which becomes the focus of our work. Prior work has applied cross-lingual transfer by using several unrelated languages as a source. Using related language can be crucial in low-resource scenarios as we discover in Section \ref{sec:ppl} and \ref{sec:asr} In our work, we limit cross-lingual transfer from one source language allowing a simpler setup for better analysis, in future, we would like to explore the impact of relatedness when the number of source languages is increased dramatically.

%% file: data.tex
\begin{table}[t!]
\footnotesize
\begin{center}
\begin{tabular}{|l|c|c|c|}
\hline \bf Language & \bf Vocabulary & \bf Train & \bf Dev\\ \hline
\multicolumn{4}{|c|}{Finnish ASR}\\
\hline
English (En) & 232K & 116M & 107K \\
Estonian (Et) & 1.7M & 97M & 33K \\
Finnish (Fi) & 1.1M & 17M & 130K \\\hline
\multicolumn{4}{|c|}{Swedish ASR}\\\hline
Danish (Da) & 2.7M & 365M & 222K \\
English (En) & 466K & 366M & 107K \\
Norwegian (No) & 2.4M & 381M & 194K\\
Swedish (Sv) & 936K & 45M & 158K\\
\hline
\end{tabular}\\
Thousands (K), Millions (M)
\end{center}
\caption{\label{tab:lm_data} The table reports the word vocabulary, training set (Train) and development set (Dev) sizes of the languages used in the experiments.}
\vspace{-0.5cm}
\end{table}
\section{Datasets}
We create two setups to evaluate cross-lingual pretraining for NNLMs. In the first setup, English (En) and Estonian (Et) are the high-resource sources of language modeling corpora, and Finnish (Fi) is the low-resource target language. In the second setup, Danish (Da), English, and Norwegian (No) are the high-resource source languages, and Swedish (Sv) is the low-resource target language.

Estonian and Finnish are contained in the Finnic language subtree, and Danish, Norwegian, and Swedish belong to the North Germanic language subtree. Thus, these source-target set of languages are considered as related languages. For both Finnish and Swedish, English, being part of the West Germanic language subtree, is considered as a more unrelated language. We also chose English as it has a large intersection for the character set, but is less mutually intelligible in comparison.

The English text is obtained from the training data of 2015 MGB Challenge \cite{bell2015mgb} consists of BBC news transcripts.
The Estonian corpus consists of web crawl text and spontaneous conversational transcripts from \newcite{MeisterEnewsci} and has been used by \newcite{8015196}.
The Finnish corpus is from Finnish Text Collection containing text from newspaper, books and novels \cite{ftc-korp_en} and has been used by \newcite{Smit2017}.
The Swedish, Danish, and Norwegian corpora, containing newspaper articles, are downloaded from Spr\aa kbanken corpus\footnote{\url{https://www.nb.no/sprakbanken}} and have been used by \newcite{swedishdata}. For Finnish as the target, more data for English was available than for Estonian, so we extract only a portion of English dataset to allow for a similar average of words per line for both datasets.  We list the corpora statistics for the various languages used in our experiments in Table \ref{tab:lm_data}.

%% file: perplexity.tex
\begin{figure}[t!]
  \centering
    \includegraphics[scale=0.4,trim={2.8cm 5.5cm 15cm 5cm},clip]{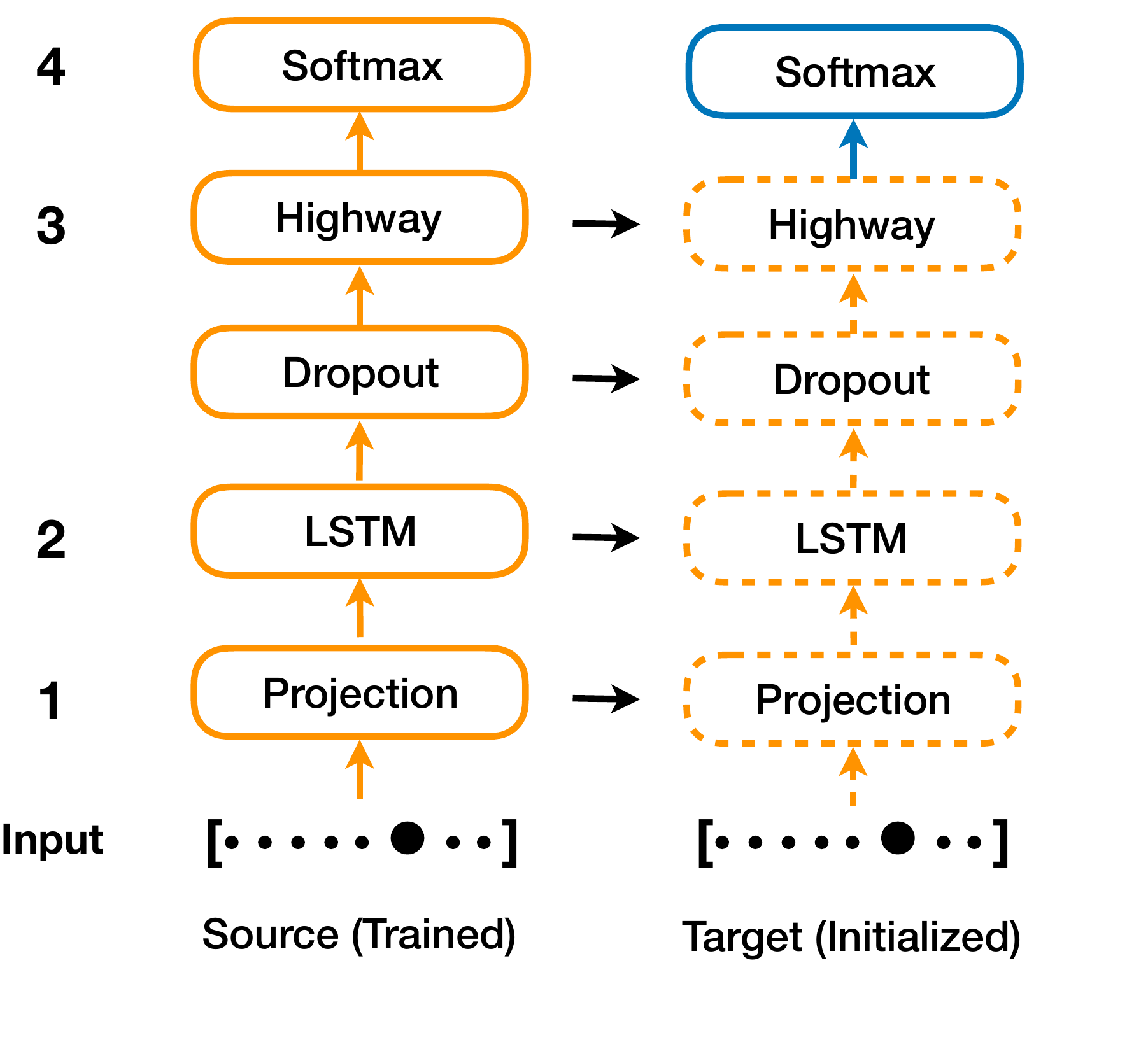}%[width=\linewidth,trim={0.1cm,0.5cm,0.1cm,0},clip]
    \vspace{-0.3cm}
    \caption{The figure displays the source and target NNLMs with the hidden layers used in our experiments. In cross-lingual pretraining, the source-language-trained hidden layers initialize parts (dotted lines) of the target-language network shown by the arrows. In contrast, the rest is randomly initialized (bold lines in the target network).}
    \vspace{-0.5cm}
    \label{fig:arch}
\end{figure}
\section{Building Language Models}
\label{sec:lms}
We train character NNLMs for our experiments and mark both the left and right ends of characters except when at the beginning or the end of a word (e.g., model = m+ +o+ +d+ +e+ +l) to achieve best results \cite{Smit2017}. With this marking scheme, we can differentiate the characters from a word into beginning (B), middle (M), end (E) and singleton units. This notation becomes relevant in Section \ref{sec:ppl}, where analyze the differences in perplexity per word position.

We build Recurrent Neural Network Language Models (RNNLM) with a projection layer (200 neurons), an LSTM layer (1000 neurons), a highway layer (1000 neurons) and a softmax output layer (displayed in Figure \ref{fig:arch}). In our experiments, both the source- and target-language neural networks have the same architecture. We train the RNNLMs using TheanoLM \cite{theanolm}, applying the adaptive gradient (Adagrad) algorithm to update the model parameters after processing a mini-batch of training examples.  The mini-batch size for models was 64, with a sequence length of 100. We used an initial learning rate of 0.1 in all the experiments and a dropout of 0.2 was used to regularize the parameter learning.
\section{Exploring Cross-Lingual Pretraining}
\label{ssec:init}
Cross-lingual pretraining involves first training the neural network on a source language. Then, starting from the input layer, the source network's hidden layers initialize the target-language neural network partially or wholly. In a partial initialization, we initialize the uninitialized layers randomly. This initialization step is followed by training on the target language, also referred to as the fine-tuning step. In both the pretraining and the fine-tuning step, the output-layer vocabulary consists of character units from all the source languages and the target language. The pretraining step transfers coarser-level information from input to higher layers into the target model and during fine-tuning, the target model refines this transferred information to a more fine-grained level.

We study neural network models across three dimensions: \textbf{1)} the source language used for pretraining step; \textbf{2)} using the number of target-model hidden layers ($l$) initialized starting from the input layer; and \textbf{3)} the amount of target language data. We represent the LM pretrained using the source language $y$ and fine-tuned using target language $z$ as $y\rightarrow z$. We vary $l$ from 1 to 4 for the architecture in Figure \ref{fig:arch}, which also shows an example for $l=3$. Here $l=1$ would refer to just initializing with the projection layer and $l=4$ would refer to initializing with all the layers. We increase the amount of target data size to match the source data size. Varying these parameters allows us to understand their effect on transfer capacity of cross-lingual pretraining.
\begin{table}[t!]
\footnotesize
\begin{center}
\begin{tabular}{|lcccc|}
\hline \multicolumn{5}{|c|}{\bf Finnish Test Set Perplexity}\\ \hline
Fi$_0$ (baseline) & \multicolumn{4}{c|}{\bf 3788}\\
\hline
$l$& 4 & 3 & 2 & 1 \\
\hline
En$\rightarrow$Fi & 4195 & 4617 & 5458 & 4211 \\
Et$\rightarrow$Fi & 3402 & 3585 & 3901 & \textbf{3009}\\\hline
\multicolumn{5}{|c|}{\bf Swedish Test Set Perplexity}\\\hline
Sv$_0$ (baseline) & \multicolumn{4}{c|}{\bf 311}\\
\hline
$l$& 4 & 3 & 2 & 1 \\
\hline
En$\rightarrow$Sv & 334 & 322 & 337 & 315 \\
No$\rightarrow$Sv & \textbf{285} & 311 & 312 & 287 \\
Da$\rightarrow$Sv & 291 & 292 & 317 & 291\\
% Da$\rightarrow$No$\rightarrow$Sv &  & 289 & 291 & 283 \\
% No$\rightarrow$Da$\rightarrow$Sv &  & 291 & 341 & \textbf{280} \\
\hline
\end{tabular}
\end{center}
\vspace{-0.3cm}
\caption{\label{tab:ppl} The table reports NNLM's test set perplexity for Finnish and Swedish using different cross-lingual initializations. For Finnish, English and Estonian are used as the source languages for pretraining. For Swedish, we use Danish, English and Norwegian as source languages. The best results in each category are marked in boldface.}
\vspace{-0.5cm}
\end{table}
\begin{figure}[t]
  \centering
    \includegraphics[height=4.8cm,trim={0.1cm 0cm 0.2cm 0},clip]{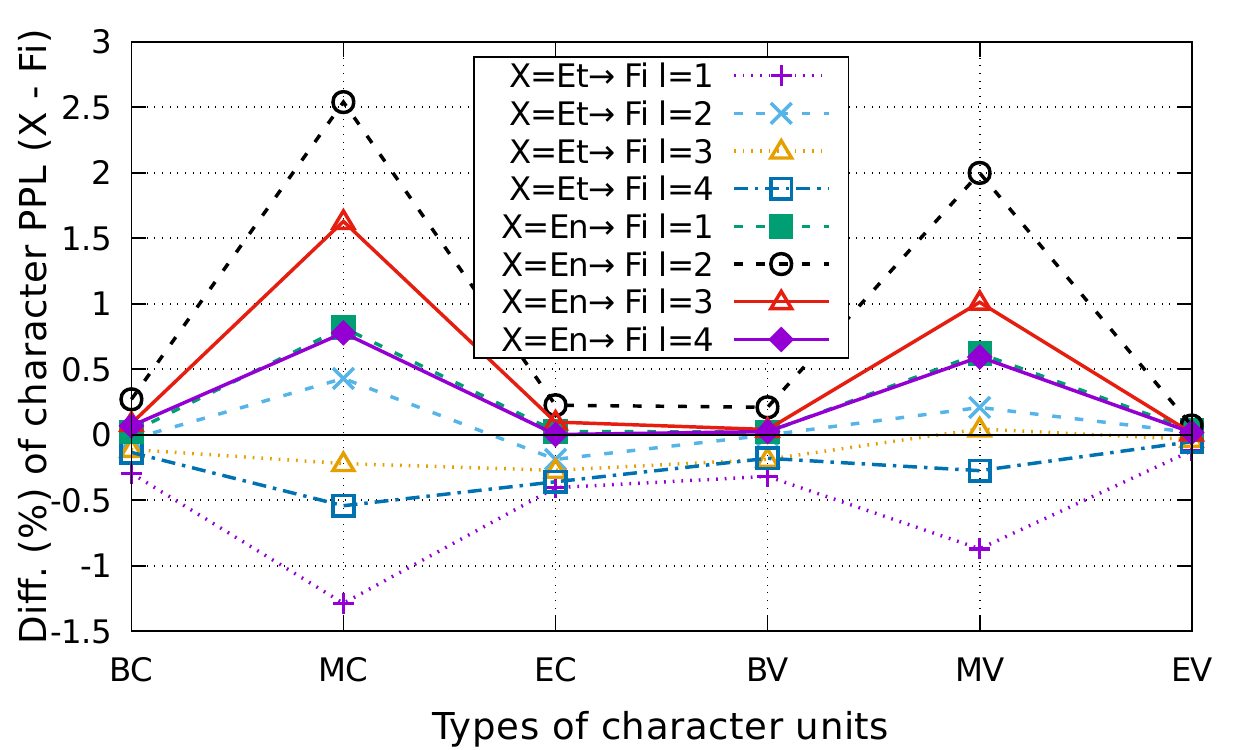}
    \vspace{-0.3cm}
    \caption{The figure shows the relative differences (\%) in character perplexity (PPL) for three different types of character units of different Finnish NNLMs on the test set. These character units exist due to the marking scheme used here: beginning (B), middle (M) and end (E), which can further be classified into consonants (C) and vowels (V).}
    \vspace{-0.5cm}
    \label{fig:fi_cppl}
\end{figure}
\section{Perplexity Experiments}
\label{sec:ppl}
Table \ref{tab:ppl} presents the test set perplexity of Finnish and Swedish LMs. When using related source languages --- like Estonian for Finnish, or Danish and Norwegian for Swedish --- to pretrain the models, we obtain better perplexity than the baseline and when English (the unrelated source language) is used, which leads to a worse perplexity for all $l$s. Using related source languages, the pretrained target LMs outperform the baseline results for most values of $l$ but, notably, when initialized with configurations $l=1,4$ of the source model. Here, we note that Finnish perplexity values are large due to long words and the domain mismatch between the training (books, newspaper articles and journals) and test (broadcast news) sets.

On characters, similar trends of perplexity improvement for related vs unrelated source language and different values of $l$ are observed. Character perplexity differences for Finnish are presented in Figure \ref{fig:fi_cppl} for different types of units, i.e., consonant (C) and vowels (V) dependent on their position in words beginning (B), middle (M) and end (E). In Figure \ref{fig:fi_cppl}, most perplexity improvements are obtained for middle consonants (MC) and middle vowels (MV), which are more frequent than other character units. For other character units, small but consistent improvements are obtained by Et$\rightarrow$Fi ($l=1$) LM over other baseline and other LMs.

For Swedish, similar improvements to Finnish results are observed for MC and MV, but some dips are seen for Danish-pretrained LMs on end consonants (EC). For brevity, we do not present this result in the paper. Overall, improvements from related-language pretraining impacts the different types of characters, enabled by a large intersection in the source-target character set.

We suspect that pretraining with a related language finds more useful information than with an unrelated one. To investigate this effect, we calculate the cosine similarity between pretrained and baseline LMs' output layer embeddings. We first find an affine transformation to align pretrained LM's embeddings with the baseline's embedding space, and then calculate the average similarity between the two sets. On Finnish, the English-pretrained embeddings have a higher average similarity (0.53) to the baseline embeddings than the Estonian-pretrained embeddings (0.51). On Swedish, similar results are observed with cosine similarity for the English-, Norwegian- and Danish-pretrained embeddings at 0.43, 0.42 and 0.42. They suggest that the related-language pretrained LMs have more conflicting information than the English-pretrained LMs. As they also perform better in terms of perplexity, the related-language pretraining seems to learn information that is complementary to the baseline LM.

%% file: asr.tex
\section{Speech Recognition Experiments}
\label{sec:asr}
For training the Finnish acoustic models, we used 1500 hours of Finnish audio from three different sources, namely, the Speecon corpus \cite{conf/lrec/IskraGMHDK02}, the Speechdat database \cite{speechdat} and the parliament corpus \cite{parl}. For testing, we used a broadcast news dataset from the Finnish national broadcaster (Yle) containing 5 hours of speech and 35k words \cite{parl}. For training Swedish acoustic models, we used 354 hours of audio provided by the Spr{\aa}kbanken corpus. From the original evaluation set, we used a total of 9 hours for development and evaluation.

The acoustic models were trained with the Kaldi toolkit \cite{Povey_ASRU2011} with a similar recipe as \cite{Smit2017}. Instead of phonemes, we use grapheme-units, as this allows for a trivial lexicon that maps between the acoustic and language modeling units. We evaluate the ASR performance in terms of Word Error Rates (WER).

For the first-pass, we train a variable-length Kneser-Ney \cite{479394} \textit{n}-gram LM using the VariKN toolkit \cite{4244538}. Then, RNNLMs, built in Section \ref{sec:lms}, are used to rescore the lattices. We also linearly interpolate cross-lingually pretrained NNLMs with target-only NNLM while optimizing the interpolation weight. We test the statistical significance of our results using the Matched Pairs Sentence Segment Word Error Test from
NIST Scoring toolkit\footnote{SCTK: \url{http://www1.icsi.berkeley.edu/Speech/docs/sctk-1.2/sctk.htm}} to compare different systems.
\begin{table}[t!]
\footnotesize
\begin{center}
\begin{tabular}{|lcccc|}
\hline \bf Language & \multicolumn{4}{c|}{\bf Baseline Architecture}\\ \hline
%En & \multicolumn{4}{c|}{31.34$^{*}$}\\
%Et & \multicolumn{4}{c|}{29.91$^{*}$}\\
Fi$_0$ (baseline) & \multicolumn{4}{c|}{\bf 16.44}\\
\hline
$l$& 4 & 3 & 2 & 1 \\
\hline
En$\rightarrow$Fi & 16.70 & 16.90$^{*}$ & 17.34$^{*}$ & 16.56 \\
Et$\rightarrow$Fi & 16.20 & 16.14$^{*}$ & 16.61 & \textbf{16.01}$^{*}$ \\
\hline
\multicolumn{5}{|c|}{\bf Linear interpolations}\\
\hline
En$\rightarrow$Fi + Fi$_0$ & 16.00$^{*}$ & 16.24$^{*}$ & 16.34 & 15.95$^{*}$ \\
Et$\rightarrow$Fi + Fi$_0$ & 15.89$^{*}$ & 15.87$^{*}$ & 16.04$^{*}$ & \textbf{15.74}$^{*}$ \\
\hline
\end{tabular}
\end{center}
% \vspace{-0.5cm}
\caption{\label{tab:fi} The table reports WER on Finnish ASR task using different cross-lingual initializations for RNNLMs used in rescoring. Here English and Estonian are used as the source languages for pretraining. Asterisks (*) denote statistical significance while comparing against Fi (16.44) using the matched pairs test with $p < 0.05$. The best results in each section are marked in boldface.}
\vspace{-0.3cm}
\end{table}
\begin{figure}[t]
  \centering
    \includegraphics[height=4.8cm,trim={0.1cm 0cm 0.2cm 0},clip]{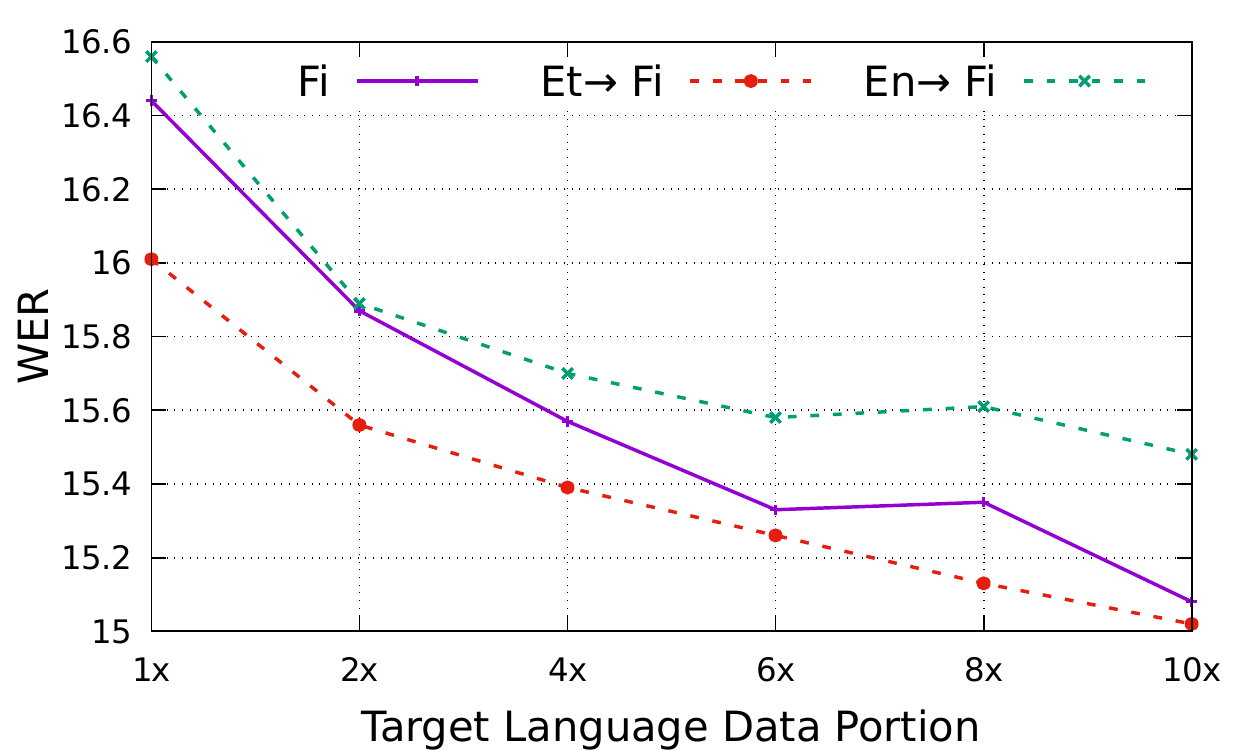}
    % \vspace{-0.3cm}
    \caption{The figures display WERs on Finnish ASR measured when varying the amount of source language data and when varying the amount of target language data.}
    % \vspace{-0.5cm}
    \label{fig:fi_wer}
\end{figure}
Tables \ref{tab:fi} and \ref{tab:sv} outline the performance of rescoring with RNNLMs (Section \ref{sec:lms}) on a Finnish and a Swedish ASR task. The first row of both these tables displays the performance of target-only trained RNNLMs (baseline). The second part reports the performance of cross-lingually pretrained models (Section \ref{ssec:init}) and the third part reports their linear interpolations with target-only baseline models.

Similar to the perplexity results (Section \ref{sec:ppl}), related source language pretraining improves the ASR performance over the baseline models and the unrelated source language pretraining degrades the performance. On Finnish ASR, English-pretrained RNNLM (En$\rightarrow$Fi) lags behind the Estonian-pretrained RNNLM (Et$\rightarrow$Fi), which also outperforms Finnish-only models. On Swedish ASR, Danish (Da$\rightarrow$Sv) and Norwegian (No$\rightarrow$Sv) pretrained models outperform the baseline and English pretrained models (En$\rightarrow$Sv). In contrast with perplexity results, lower-layer ($l=1$) based initialization shows the most benefit over the higher-layer ($l=2,3,4$) initializations for both Finnish and Swedish ASR. We note that quite like perplexity results, ASR performance on Swedish is lower than Finnish as the Swedish task is easier than the Finnish one.

\begin{table}[!t]
\footnotesize
\begin{center}
\begin{tabular}{|lcccc|}
\hline \bf Language & \multicolumn{4}{c|}{\bf Baseline Architecture}\\ \hline
%En & \multicolumn{2}{c|}{13.78$^{*}$}\\
%No & \multicolumn{2}{c|}{12.52$^{*}$}\\
%Da & \multicolumn{2}{c|}{12.24$^{*}$}\\
Sv$_0$ (baseline) & \multicolumn{4}{c|}{\bf 4.42}\\
\hline
$l$ & 4 & 3 & 2 & 1 \\
\hline
En $\rightarrow$ Sv & 4.43 & 4.46 & 4.62 & 4.41 \\
No $\rightarrow$ Sv & 4.18 & 4.42 & 4.38 & 4.17$^{*}$ \\
Da $\rightarrow$ Sv & 4.24 & 4.16$^{*}$ & 4.39 & \textbf{4.15$^{*}$} \\
% Da $\rightarrow$ No $\rightarrow$ Sv & & 4.13$^{*}$ & 4.68 & 4.16$^{*}$ \\
%No $\rightarrow$ Da $\rightarrow$ Sv & & 4.20$^{*}$ & 4.17 & \textbf{4.10}$^{*}$\\
\hline
\multicolumn{5}{|c|}{\bf Linear Interpolations}\\
\hline
En $\rightarrow$ Sv + Sv$_0$ & 4.15$^{*}$ & 4.18$^{*}$ & 4.20 & 4.15$^{*}$ \\
No $\rightarrow$ Sv + Sv$_0$ & 4.02$^{*}$ & 4.40 & 4.35 & 4.00$^{*}$ \\
Da $\rightarrow$ Sv + Sv$_0$ & 4.01$^{*}$ & 4.04$^{*}$ & 4.15$^{*}$ &\textbf{3.98$^{*}$} \\
%Da $\rightarrow$ No $\rightarrow$ Sv + Sv & & 4.02$^{*}$ & 4.03$^{*}$ & 3.99$^{*}$ \\
%No $\rightarrow$ Da $\rightarrow$ Sv + Sv & & 4.03$^{*}$ & 4.26 & \textbf{3.95}$^{*}$\\
\hline
\end{tabular}
\end{center}
% \vspace{-0.5cm}
\caption{\label{tab:sv} The table reports WER on Swedish ASR task for different configurations of RNNLMs used in rescoring. Here Danish, English and Norwegian are used as the source languages for cross-lingual pretraining. Asterisks (*) denote statistical significance when comparing to Sv (4.41) using the matched pairs test with $p < 0.05$. The best results in each section are marked in boldface.}
\vspace{-0.5cm}
\end{table}

In Figure \ref{fig:fi_wer}, we observe little performance increase by cross-lingual pretraining when we vary the target data size by increasing it to comparable sizes of source language data. At least for Estonian, increasing Finnish data (target) closes the gap between cross-lingual pretraining and target-only model. The cross-lingual transfer seems to work best with a larger number of resources for the related source language in comparison to the target language.

Furthermore, interpolations of the baseline model with the cross-lingually pretrained models improve over its constituent models. On both Finnish and Swedish ASR, cross-lingual pretraining with English combined with the baseline model can outperform the baseline model, unlike when used individually. This improvement can be attributed to the regularization effect of such an interpolation. Linear interpolations based on other source languages like Estonian, Danish and Norwegian further improve the results consistently across different initialization schemes. We hypothesize that this effect is due to the complementary information learned by these related-language models. Overall, the individual systems and the interpolations based on related source languages show a significant and the most substantial improvement in performance.

%% file: conclusion.tex
\section{Concluding Remarks}
We investigated cross-lingual transfer for character-based neural network language models in a low-resource scenario. Cross-lingual pretraining with related source language significantly improved (3-6\% relative) over no pretraining, whereas pretraining with unrelated source language had adverse effects. At a character level, we suspect cross-lingual pretraining works for related languages as they share a large portion of the character set. The large shared vocabulary provides soft alignments between characters in related languages supporting the transfer of relevant information from source to target models. This information transfer is in contrast to multi-character units where the transfer is dependent on shared anchor tokens (like numbers, proper nouns). However, we still lack an empirical understanding of this phenomenon and in our future work, we hope to explore this phenomenon.

Additionally, transferring the lower layer information and having more source data than target data was significant for low-resource ASR. As a followup to our study, we investigate the effects of language relatedness for cross-lingual pretraining in transformer-based language models.

%% file: lrec2020W-xample-kc.bbl
\begin{thebibliography}{}

\bibitem[\protect\citename{Aharoni \bgroup et al.\egroup
  }2019]{aharoni-etal-2019-massively}
Aharoni, R., Johnson, M., and Firat, O.
\newblock (2019).
\newblock Massively multilingual neural machine translation.
\newblock In {\em Proceedings of the 2019 Conference of the North {A}merican
  Chapter of the Association for Computational Linguistics: Human Language
  Technologies, Volume 1 (Long and Short Papers)}, pages 3874--3884,
  Minneapolis, Minnesota, June. Association for Computational Linguistics.

\bibitem[\protect\citename{Bell \bgroup et al.\egroup }2015]{bell2015mgb}
Bell, P., Gales, M.~J., Hain, T., Kilgour, J., Lanchantin, P., Liu, X.,
  McParland, A., Renals, S., Saz, O., Wester, M., et~al.
\newblock (2015).
\newblock The mgb challenge: Evaluating multi-genre broadcast media
  recognition.
\newblock In {\em 2015 IEEE Workshop on Automatic Speech Recognition and
  Understanding (ASRU)}, pages 687--693.

\bibitem[\protect\citename{Conneau and Lample}2019]{DBLP:conf/nips/ConneauL19}
Conneau, A. and Lample, G.
\newblock (2019).
\newblock Cross-lingual language model pretraining.
\newblock In {\em Advances in Neural Information Processing Systems 32: Annual
  Conference on Neural Information Processing Systems 2019, NeurIPS 2019, 8-14
  December 2019, Vancouver, BC, Canada}, pages 7057--7067.

\bibitem[\protect\citename{Conneau \bgroup et al.\egroup
  }2019]{DBLP:journals/corr/abs-1911-02116}
Conneau, A., Khandelwal, K., Goyal, N., Chaudhary, V., Wenzek, G.,
  Guzm{\'{a}}n, F., Grave, E., Ott, M., Zettlemoyer, L., and Stoyanov, V.
\newblock (2019).
\newblock Unsupervised cross-lingual representation learning at scale.
\newblock {\em CoRR}, abs/1911.02116.

\bibitem[\protect\citename{{CSC - IT Center for Science}}1998]{ftc-korp_en}
{CSC - IT Center for Science}.
\newblock (1998).
\newblock {The Helsinki Korp Version of the Finnish Text Collection}.

\bibitem[\protect\citename{Enarvi and Kurimo}2016]{theanolm}
Enarvi, S. and Kurimo, M.
\newblock (2016).
\newblock Theanolm - an extensible toolkit for neural network language
  modeling.
\newblock In {\em {INTERSPEECH}}, pages 5; 3052--3056.

\bibitem[\protect\citename{{Enarvi} \bgroup et al.\egroup }2017]{8015196}
{Enarvi}, S., {Smit}, P., {Virpioja}, S., and {Kurimo}, M.
\newblock (2017).
\newblock Automatic speech recognition with very large conversational {Finnish}
  and {Estonian} vocabularies.
\newblock {\em IEEE/ACM Transactions on Audio, Speech, and Language
  Processing}, 25(11):2085--2097.

\bibitem[\protect\citename{Goodman}2001]{DBLP:conf/icassp/Goodman01}
Goodman, J.
\newblock (2001).
\newblock Classes for fast maximum entropy training.
\newblock In {\em {ICASSP}}, pages 561--564.

\bibitem[\protect\citename{Iskra \bgroup et al.\egroup
  }2002]{conf/lrec/IskraGMHDK02}
Iskra, D.~J., Grosskopf, B., Marasek, K., van~den Heuvel, H., Diehl, F., and
  Kießling, A.
\newblock (2002).
\newblock Speecon - speech databases for consumer devices: Database
  specification and validation.
\newblock In {\em LREC}. European Language Resources Association.

\bibitem[\protect\citename{Kim \bgroup et al.\egroup
  }2019]{DBLP:conf/acl/KimGN19}
Kim, Y., Gao, Y., and Ney, H.
\newblock (2019).
\newblock Effective cross-lingual transfer of neural machine translation models
  without shared vocabularies.
\newblock In {\em Proceedings of the 57th Conference of the Association for
  Computational Linguistics, {ACL} 2019, Florence, Italy, July 28- August 2,
  2019, Volume 1: Long Papers}, pages 1246--1257. Association for Computational
  Linguistics.

\bibitem[\protect\citename{Kneser and Ney}1995]{479394}
Kneser, R. and Ney, H.
\newblock (1995).
\newblock Improved backing-off for m-gram language modeling.
\newblock In {\em {ICASSP}}, volume~1, pages 181--184.

\bibitem[\protect\citename{Le \bgroup et al.\egroup
  }2011]{DBLP:conf/interspeech/LeOMAGY11}
Le, H.~S., Oparin, I., Messaoudi, A.~K., Allauzen, A., Gauvain, J.~L., and
  Yvon, F.
\newblock (2011).
\newblock Large vocabulary {SOUL} neural network language models.
\newblock In {\em {INTERSPEECH}}, pages 1469--1472.

\bibitem[\protect\citename{Mansikkaniemi \bgroup et al.\egroup }2017]{parl}
Mansikkaniemi, A., Smit, P., and Kurimo, M.
\newblock (2017).
\newblock Automatic construction of the {Finnish} parliament speech corpus.
\newblock In {\em INTERSPEECH}, pages 3762--3766.

\bibitem[\protect\citename{Meister \bgroup et al.\egroup }2012]{MeisterEnewsci}
Meister, E., Meister, L., and Metsvahi, R.
\newblock (2012).
\newblock New speech corpora at {IoC}.
\newblock In {\em {XXVII} Fonetiikan p{\"a}iv{\"a} --- Phonetics Symposium},
  pages 30--33.

\bibitem[\protect\citename{Morin and Bengio}2005]{DBLP:conf/aistats/MorinB05}
Morin, F. and Bengio, Y.
\newblock (2005).
\newblock Hierarchical probabilistic neural network language model.
\newblock In {\em {AISTATS}}.

\bibitem[\protect\citename{Povey \bgroup et al.\egroup }2011]{Povey_ASRU2011}
Povey, D., Ghoshal, A., Boulianne, G., Burget, L., Glembek, O., Goel, N.,
  Hannemann, M., Motlicek, P., Qian, Y., Schwarz, P., Silovsky, J., Stemmer,
  G., and Vesely, K.
\newblock (2011).
\newblock The kaldi speech recognition toolkit.
\newblock In {\em ASRU}. IEEE Signal Processing Society, December.

\bibitem[\protect\citename{Rosti \bgroup et al.\egroup }1998]{speechdat}
Rosti, A., {R\"am\"o}, A., Saarelainen, T., and Yli-Hietanen, J.
\newblock (1998).
\newblock Speechdat {Finnish} database for the fixed telephone network.
\newblock Technical report, Tampere University of Technology.

\bibitem[\protect\citename{Siivola \bgroup et al.\egroup }2007]{4244538}
Siivola, V., Hirsimaki, T., and Virpioja, S.
\newblock (2007).
\newblock On growing and pruning {Kneser-Ney} smoothed $ n$-gram models.
\newblock {\em IEEE Transactions on Audio, Speech, and Language Processing},
  15(5):1617--1624.

\bibitem[\protect\citename{Smit \bgroup et al.\egroup }2017]{Smit2017}
Smit, P., Gangireddy, S.~R., Enarvi, S., Virpioja, S., and Kurimo, M.
\newblock (2017).
\newblock Character-based units for unlimited vocabulary continuous speech
  recognition.
\newblock In {\em ASRU}, pages 149--156.

\bibitem[\protect\citename{Smit \bgroup et al.\egroup }2018]{swedishdata}
Smit, P., Virpioja, S., and Kurimo, M.
\newblock (2018).
\newblock Advances in subword-based hmm-dnn speech recognition across
  languages.
\newblock Technical report, Aalto University.

\bibitem[\protect\citename{Zhuang \bgroup et al.\egroup
  }2017]{DBLP:conf/interspeech/ZhuangGRPL17}
Zhuang, X., Ghoshal, A., Rosti, A., Paulik, M., and Liu, D.
\newblock (2017).
\newblock Improving {DNN} bluetooth narrowband acoustic models by
  cross-bandwidth and cross-lingual initialization.
\newblock In {\em {INTERSPEECH}}, pages 2148--2152.

\end{thebibliography}
